# AN EVALUATION OF TWO ALTERNATIVES TO MINIMAX

Dana Nau[1]
Computer Science Department
University of Maryland
College Park, MD 20742

Paul Purdom
Computer Science Department
Indiana University
Bloomington, IN 47405

Chun-Hung Tzeng
Computer Science Department
Ball State University
Muncie, IN 47306

## Abstract

In the field of Artificial Intelligence, traditional approaches to choosing moves in games involve the use of the minimax algorithm. However, recent research results indicate that minimaxing may not always be the best approach. In this paper we summarize the results of some measurements on several model games with several different evaluation functions. These measurements, which are presented in detail in [NPT], show that there are some new algorithms that can make significantly better use of evaluation function values than the minimax algorithm does.

## 1. Introduction

This paper is concerned with how to make the best use of evaluation function values to choose moves in games and game trees. The traditional approach used in Artificial Intelligence is to combine the values using the minimax algorithm. Previous work by Nau [Na83b, Na82], Pearl [Pe82], and Tzeng and Purdom [TP, Tz] has shown that this approach may not always be best. The current paper summarizes the results of a study involving measurements on several model games with several different evaluation functions and several different ways of combining the evaluation function values. These measurements show that there are some new algorithms that for some games can make significantly better use of evaluation function values than the minimax algorithm does. These results are discussed in detail in [NPT].

Three methods of propagating the estimates from evaluation function are compared in this paper: *minimax propagation* (which is well-known [Ni]),[2] *product propagation* (which treats the evaluation function values as if they were independent probabilities [Na83a]), and a decision rule which is intermediate between these two, which for this paper we call *average propagation*.

Minimax propagation is the best way to combine values if one's opinions of the values of previously analyzed positions will not change on later moves. However, real game playing programs reanalyze positions after each move is made, and usually come up with slightly different opinions on the later analyses (because, as the program gets closer to a position, it is able to search more levels past the position). (Minimax propagation is also known to be the best way to combine values at a node $N$ if those values are the exact values. But if one can obtain exact values, then there is no need for searching at all, and thus no need for combining values.)

Product propagation is the best way to combine values if they are estimates of (independent) probabilities of forced wins and if no one is going to make any mistakes after the first move. But using estimates (which contain errors) of position values on the first move and then making perfect moves for the rest of the game is equivalent to using an estimator with errors for the first move and a perfect estimator for later moves. This implies a drastic reevaluation of the positions after the first move is made.

---

[1] This work was supported in part by an NSF Presidential Young Investigator award to Dana Nau, including matching funds from IBM, Martin Marietta, and General Motors.

---

[2] Decision analysts refer to minimax propagation as the maximin decision criterion.

The situation encountered in real game playing is generally somewhere between the two extremes described above. If a game playing program eventually moves to some node $N$, then the values computed at each move in the game are progressively more accurate estimates of the value of $N$. Although the errors in these estimates decrease after each move, they usually do not drop to zero. Therefore, it should be better to use an approach which is intermediate between the two extremes of minimax propagation and product propagation. There are many possible propagation methods satisfying this requirement, and we chose to study one (namely average propagation) whose values are easy to calculate.

We compared the three propagation rules on several related classes of two-person board-splitting games, using several evaluation functions:

(1) P-games (as defined in [Na82a]) using an evaluation function $e_1$ described in [Na82a];

(2) P-games using an evaluation function $e_2$ which computes the exact probability that a position in a P-game is a forced win, given various relevant features of the position;

(3) N-games (as defined in [Na82a]) using $e_1$;

(4) G-games (as defined in [Na83c]) using $e_1$;

(5) G-games using an evaluation function $e_3$ particularly suited for G-games.

## 2. Results and Data Analysis

It is difficult to conclude much about any propagation methods by considering how it does on a single game. One cannot tell from a single trial whether a method was good or merely lucky. Therefore, each comparison was done on a large set of games.

Comparisons (1), (2), and (3) were done using 1600 randomly generated pairs of games, each chosen in such a way that the game would be ten moves long. Each pair of games was played on a single game board; one game was played with one player moving first and another was played with his opponent moving first. For each pair of games we had 10 contests, one for each depth of searching from 1 to 10. Each contest included all 1600 pairs of games. Most game boards were such that the starting position (first player to move or second player to move) rather than the propagation method determined who won the game, but for some game boards one propagation method was able to win both games of the pair. We call these latter games *critical games*.

The comparisons showed that for the set of games considered, average propagation was always as good as and often several percent better than either minimax propagation or product propagation. Product propagation was usually better than minimax propagation, but not at all search depths.

An important question is how significant the results are. Even if two methods are equally good on the average, chance fluctuations would usually result in one of the methods winning over half the games in a 1600 game contest. To test the significance of each result, we consider the null hypothesis that the number of pairs of wins (among the critical games) was a random event with probability 1/2. If the significance level (the probability that the observed deviation from 1/2 could have arisen by chance) is below, say, 5%, then we say that the method that won over 50% of the games in this sample performed significantly better than its opponent.

The results of comparison (1) are shown in Tables 1 and 2.[3] In this comparison, product propagation did significantly better than minimax propagation at most search depths. Minimax propagation was better for search depth 3. For depths 2 and 5, the results were too close to be sure which method was better. For depths 3, 4, 6, 7, and 8 product propaga-

---

[3] Space limitations do not permit the inclusion of tables for any comparisons other than comparison (1). For tables showing the details of the other comparisons, the reader is referred to [NPT].

**Table 1.**—Number of pairs of P-games won by (1) product propagation against minimax propagation, (2) average propagation against minimax propagation, and (3) average propagation against product propagation, with both players searching to the same depth $d$ using the evaluation function $e_1$. The results come from Monte Carlo simulations of 1600 game boards each. For each game board and each value of $d$, a pair of games was played, so that each player had a chance to start first. All players were using the same evaluation function $e_1$. Out of the 1600 pairs, a pair was counted only if the same player won both games in the pair.

| Search depth | Product vs. Minimax | | Average vs. Minimax | | Average vs. Product | | Notes |
|---|---|---|---|---|---|---|---|
| | Number of pairs | Number of wins | Number of pairs | Number of wins | Number of pairs | Number of wins | |
| 1 | 0 | 0 | 0 | 0 | 0 | 0 | * |
| 2 | 472 | 231 | 320 | 181 | 240 | 140 | |
| 3 | 569 | 249 | 411 | 218 | 332 | 199 | |
| 4 | 597 | 334 | 520 | 331 | 352 | 221 | |
| 5 | 577 | 290 | 478 | 308 | 341 | 227 | |
| 6 | 567 | 348 | 525 | 385 | 266 | 191 | |
| 7 | 424 | 235 | 352 | 229 | 205 | 140 | |
| 8 | 324 | 223 | 305 | 236 | 95 | 70 | |
| 9 | 0 | 0 | 0 | 0 | 0 | 0 | *, ** |
| 10 | 0 | 0 | 0 | 0 | 0 | 0 | *, ** |

\* For search depths 1, 9, and 10, both players play identically.
\*\* For search depths 9 and 10, both players play perfectly.

**Table 2.**—Percentage of pairs of P-games won by (1) product propagation against minimax propagation, (2) average propagation against minimax propagation, and (3) average propagation against product propagation, with both players searching to the same depth $d$ using the evaluation function $e_1$. The data is from the same games used for Table 1. The significance column gives the probability that the data is consistent with the null hypothesis that each method is equally good. Small numbers (below 5%, for example), indicate that the deviation in the number of wins from 50% is unlikely to be from a chance fluctuations, while large numbers indicate that from this data one cannot reliably conclude which method is best.

| Search depth | Product vs. Minimax | | Average vs. Minimax | | Average vs. Product | |
|---|---|---|---|---|---|---|
| | Wins | Significance | Wins | Significance | Wins | Significance |
| 2 | 48.9% | 65.% | 56.6% | 1.9% | 58.3% | 1.2% |
| 3 | 43.8% | 0.28% | 53.0% | 23.% | 59.9% | $3 \times 10^{-2}$% |
| 4 | 55.9% | 0.38% | 63.7% | $1 \times 10^{-7}$% | 62.8% | $2 \times 10^{-4}$% |
| 5 | 50.3% | 90.% | 64.4% | $2 \times 10^{-8}$% | 66.6% | $9 \times 10^{-8}$% |
| 6 | 61.4% | $6 \times 10^{-6}$% | 73.3% | $1 \times 10^{-24}$% | 71.8% | $1 \times 10^{-10}$% |
| 7 | 55.4% | 2.6% | 65.1% | $2 \times 10^{-6}$% | 68.3% | $2 \times 10^{-5}$% |
| 8 | 68.8% | $1 \times 10^{-9}$% | 77.4% | $1 \times 10^{-19}$% | 73.7% | $4 \times 10^{-4}$% |

tion clearly did better.[4]

Comparison (1) also showed average propagation to be a clear winner over minimax propagation in P-games when $e_1$ is used. Only at depth 3 were the results close enough for there to be any doubt. In addition, average propagation was a clear winner over product propagation at all search depths.

There are theoretical reasons to believe that product propagation should do even better on P-games when $e_2$ is used rather than $e_1$ [TP], and the results of comparison (2) corroborated this. In comparison (2), average propagation and product propagation both did better in comparison to minimax propagation than they had done before: for search depths 4, 5, 6, 7, and 8, the significance levels were all at $10^{-8}\%$ or better.[5] In comparison (2), average propagation appeared to do better than product propagation at most search depths, but the results were not statistically significant except at search depth 4, where they were marginally significant. These results show that product propagation becomes relatively better compared to both minimax propagation and average propagation when better estimates are used for the probability that a node is a forced win.

The results of comparison (3) suggest that for this set of games average propagation may again be the best, but the differences among the methods are much smaller. This time minimax propagation is better than product propagation for search depths 3 and 4 (and probably 2). Average propagation may be better than minimax propagation at larger search depths (all the results were above 50%), but one cannot be sure based on this data, because the significance levels were all above 20%. Average propagation is significantly better than product propagation for all search depths except 8, where the results are inconclusive. It is more difficult to draw definite conclusions for N-games partly

---

[4] Search depths 1, 9, and 10 are irrelevant in this comparison, because at search depth 1, all three propagation rules choose exactly the same moves, and at depths 9 and 10 the evaluation function yields perfect play.

[5] Search depths 1, 9, and 10 are irrelevant in this comparison for the same reasons as in comparison (1).

because there is a low percentage of critical games.

There are only 2048 initial playing boards for G-games of ten moves, so for comparisons (4) and (5) it was possible to enumerate all these boards and obtain exact values rather than Monte Carlo estimates. In comparison (4), product propagation and average propagation both did somewhat better than minimax propagation, and did about the same as each other. In comparison (5), average propagation and product propagation still did about equally well, but this time both did somewhat worse than minimax propagation. One possible reason for this is discussed in [NPT].

### 3. Conclusion

The main conclusions of this study are that the method used to back up estimates has a definite effect on the quality of play, and that the traditional minimax propagation method not always the best method to use. Which method of propagation works best depends on both the estimator and the game.

Some of our students are extending these investigations to games that are more commonly known. Teague [Te] has shown that minimax propagation does markedly better than product propagation and average propagation in the game of Othello, but Chi [Ch] has preliminary results which appear to indicate that both product propagation and average propagation outperform minimax propagation in a modified version of Kalah.

One problem with methods other than minimax propagation is that the value of every node has some effect on the final result. Thus methods such as the alpha-beta pruning procedure cannot be used to speed up the search without affecting the final value computed. Programs for most games use deep searches, and these programs will not be able to make much use of these new methods unless suitable pruning procedures are found. A method is needed which will always expand the node that is expected to have the largest effect on the value.

The games where the new results may have the most immediate application are probabilistic games such as backgammon, where it is not feasible to do deep searches of the game tree. Since alpha-beta pruning does not save significant amounts of work on shallow searches, it is conceivable that such games can profit immediately from improved methods of backing up values.